\documentclass{article}
\usepackage{spconf,amsmath,graphicx,tabularx,booktabs,wasysym,xcolor}
\usepackage[skip=0pt]{caption}
\pdfoutput=1

\title{END-TO-END DIAGNOSIS AND SEGMENTATION LEARNING \\FROM CARDIAC MAGNETIC RESONANCE IMAGING}

\name{
    Gerard Snaauw$^{a,b}$ \qquad
    Dong Gong$^{a}$ \qquad
    Gabriel Maicas$^{a}$ \qquad
    }
\secondlinename{
    Anton van den Hengel$^{a}$ \qquad
    Wiro J. Niessen$^{b,c}$ \qquad
    Johan Verjans$^{a,d}$ \qquad
    Gustavo Carneiro$^{a}$ 
}

\address{
$^{a}$ AIML, School of Computer Science, University of Adelaide, Australia \\
$^{b}$ Imaging Physics, Faculty of Applied Sciences, Delft University of Technology, Netherlands \\
$^{c}$ Department of Radiology \& Medical Informatics, Erasmus MC, Rotterdam, Netherlands \\
$^{d}$ South Australian Health and Medical Research Institute, Adelaide, Australia
}

\begin{document}
\ninept
\maketitle
\begin{abstract}
Cardiac magnetic resonance (CMR) is used extensively in the diagnosis and management of cardiovascular disease. Deep learning methods have proven to deliver segmentation results comparable to human experts in CMR imaging, but there have been no convincing results for the problem of end-to-end segmentation and diagnosis from CMR. This is in part due to a lack of sufficiently large datasets required to train robust diagnosis models. 
In this paper, we propose a learning method to train diagnosis models, where our approach is designed to work with relatively small datasets.  In particular, the optimisation loss is based on multi-task learning that jointly trains for the tasks of segmentation and diagnosis classification.
We hypothesize that segmentation has a regularizing effect on the learning of features relevant for diagnosis.  
Using the 100 training and 50 testing samples available from the Automated Cardiac Diagnosis Challenge (ACDC) dataset, which has a balanced distribution of 5 cardiac diagnoses, we observe a reduction of the classification error from 32\% to 22\%, and a faster convergence compared to a baseline without segmentation.  To the best of our knowledge, this is the best diagnosis results from CMR using an end-to-end diagnosis and segmentation learning method.
\end{abstract}
\vspace{-.05in}
\section{Introduction} 
\label{sec:intro}
\vspace{-.05in}

Cardiovascular disease (CVD) is consistently ranked the leading cause of death worldwide, killing more people in 2016 than the next four causes together~\cite{WHO}. 
Cardiovascular Magnetic Resonance (CMR) imaging has proven to be of great value in CVD diagnosis and management. 
A combination of factors such as lack of ionizing radiation, excellent soft tissue contrast, and high reproducibility have made it the preferred imaging modality in the quantification of ventricular volumes, myocardial function and scarring visualization~\cite{Salerno2017,Peng2016}. 
Increasing clinical use has also resulted in an increased application of CMR in large cohort studies~\cite{Medrano2015}. 
This proliferation of medical imaging datasets will impact the need for automated tools, making machine learning for imaging data a very promising field.

Current machine learning based methods for automated cardiac diagnosis focus on the detection and segmentation of the heart, followed by the extraction of handcrafted features that are then used for diagnosis~\cite{Bernard2018}. This approach is reflected in the 2017 Automated Cardiac Diagnosis Challenge (ACDC) where the aim is to automatically perform segmentation and diagnosis on a 4D cine-CMR scan. All but one participant in the segmentation part of the challenge used deep learning, where structures like U-net and dilated convolutional network were explored -- the best deep learning approaches scored on par with clinical experts. Interestingly enough, none of the participants in the diagnosis part of the challenge used deep learning. Instead, they performed classification using support vector machines (SVM) and random forests (RF) on handcrafted features extracted from segmentation maps~\cite{Bernard2018}.

The design and implementation of handcrafted features have numerous disadvantages~\cite{Litjens2017}: sub-optimality for the classification task, requirement of a manual re-design process for new tasks, in-depth knowledge of the task for the design of relevant features, etc.
One of the major motivations for the development of deep learning models is exactly the automatic design of features that are learned to solve particular classification tasks -- this mitigates all the negative points listed above.
In fact, deep learning has consistently shown state-of-the-art segmentation and classification results~\cite{He2018}. However, to our knowledge, there have been no convincing attempt at an end-to-end learning for segmentation and diagnosis classification in cardiology. One possible explanation for this is the lack of large datasets available for this task~\cite{Litjens2017}.

In this paper, we propose a multi-task learning process that combines cardiac segmentation and diagnosis classification using the ACDC dataset, which has a balanced distribution of 5 cardiac diagnoses.  
This multi-task learning guides the automatic design of features relevant for both tasks and serves two purposes: 1) regularization of cardiac diagnosis training process, and 2) reduction of convergence time. 
In addition to this, we evaluate an exponential version of the linear Dice loss to overcome the current limitations of segmenting objects with different sizes, as inspired by Wong et al.~\cite{Wong2018}. Results show a reduction of the classification error from 32\% to 22\%, and a faster convergence compared to a baseline without segmentation.  To the best of our knowledge, this is the best result of an end-to-end trained segmentation and classification method for diagnosing from CMR.

\vspace{-.05in}

\section{Materials and Methods} \label{sec:methods}
\vspace{-.05in}

\subsection{Dataset} 
\label{ssec:dataset}

The ACDC dataset consists of training and testing sets with 100 and 50 4D cine-CMR scans, respectively. Both sets contain a balanced distribution of the following five classes: \{Normal (NOR), dilated cardiomyopathy (DCM), hypertropic cardiomyopathy (HCM), prior myocardial infarction (MINF), abnormal right ventricle (ARV)\}.  These sets also contain manual segmentation of the \{left ventricular cavity (LV), right ventricular cavity (RV), LV myocardium (Myo), background (BG)\} at the end-systolic (ES) and end-diastolic (ED) phases.

Scans are resampled to $1.0 \times 1.0$ mm in the in-slice plane; then they are center cropped and normalized to zero mean and standard deviation one. Center cropping is performed around the heart bounding box, which is extracted from the segmentation maps for the training set and defined manually for the test set. 
Normalization is performed slice by slice since cine scans are acquired in this manner. The volumes from ED and ES phases are combined to form a triple-channel input (ED, S, ES) for the network, where S $=$ ED $-$ ES represents the subtraction volume of the two phases, explicitly incorporating temporal information.

\subsection{Network Architecture} 
\label{ssec:network}

\begin{figure}[t]
  \centering
  \centerline{\includegraphics[width=8.4cm]{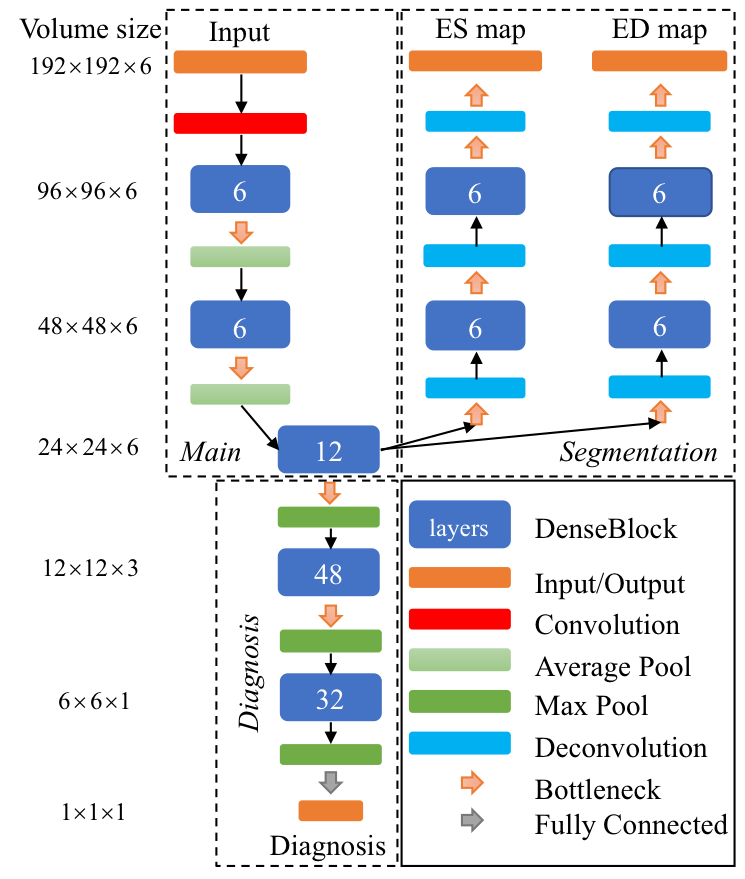}}
\caption{
Network architecture consists of three branches. The shared and diagnosis branches form a DenseNet structure while shared and segmentation form a U-net like structure.
Six consecutive slices of both phases and their subtraction volume are combined in a three channel input (ED, ED-ES, ES) to include phase information.
} \label{fig:network}
\end{figure}

The DenseNet~\cite{Huang2016} and U-net~\cite{Ronneberger2015} models are combined to solve the classification and segmentation tasks. Three distinct branches are identified in the network, namely (Fig.~\ref{fig:network}): main (MB), segmentation (SB) and diagnosis (DB) branches.
The composite function for every operation in the model consists of \emph{Operation-BN-ReLU}, with BN denoting batch normalization, and ReLU, rectified linear units. MB applies a 7x7x7 convolution with stride 2 in the in-slice direction of the input to generate the initial 64 feature maps. Hereafter, the model follows a \emph{DenseBlock-bottleneck-size manipulation} structure.
DenseBlocks consist of 3x3x3 convolution layers with growth rate $k=12$ and an increasing number of layers per block as feature map sizes get smaller. 
Bottleneck layers apply 1x1x1 convolutions to halve the number of feature maps. 
Size manipulation layers halve or double feature map sizes depending on the branch. MB applies average pooling to produce optimal features for both tasks, DB applies max pooling to learn optimal features for classification, and SB applies transpose convolutions to produce segmentation masks at original input size. MB and SB manipulate feature map sizes in the in-plane direction while DB manipulates all three axes.

\subsection{Loss functions} 
\label{ssec:loss}
The network is trained using a loss function consisting of a convex combination of a diagnosis classification and segmentation losses:
\begin{equation} \label{eq:loss}
    \mathcal{L}=\alpha\cdot\mathcal{L}_{Diagnosis} +(1-\alpha)\cdot\mathcal{L}_{Segmentation},
\end{equation}
where the diagnosis loss is evaluated using the standard cross-entropy loss, and the segmentation loss is represented by:
\begin{equation} \label{eq:seg_loss}
    \displaystyle \mathcal{L}_{Segmentation}= \frac{1}{N} \sum_{n=1}^N (1-Dice_n)^p,
\end{equation}
with $N=4$ denoting the number of segmentation labels in the dataset and $p$ being a parameter to control the shape of the loss function. If $p=1$, $(1-Dice)^p$ becomes the binary Dice loss~\cite{Milletari2016}. This linear Dice loss is known to produce less accurate segmentation results in unbalanced datasets, i.e., containing objects of different sizes. To overcome this limitation, we evaluate the loss with different values for $p$. 

\begin{figure*}[t]
    \begin{minipage}[b]{0.4\linewidth}
      \centering
      \centerline{\includegraphics[width=8cm]{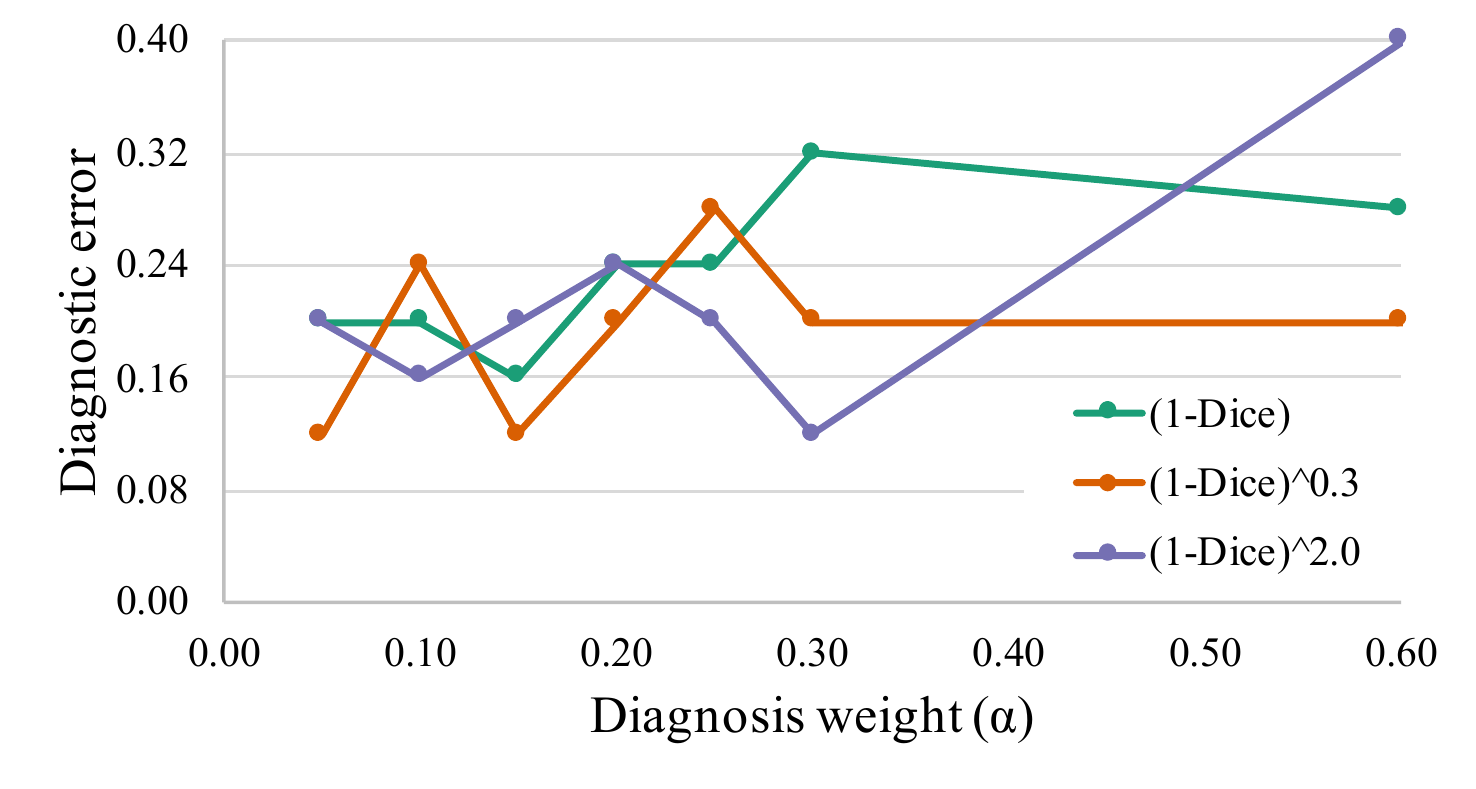}}
    \end{minipage}
    \hfill
    \begin{minipage}[b]{.4\linewidth}
      \centering
      \centerline{\includegraphics[width=8cm]{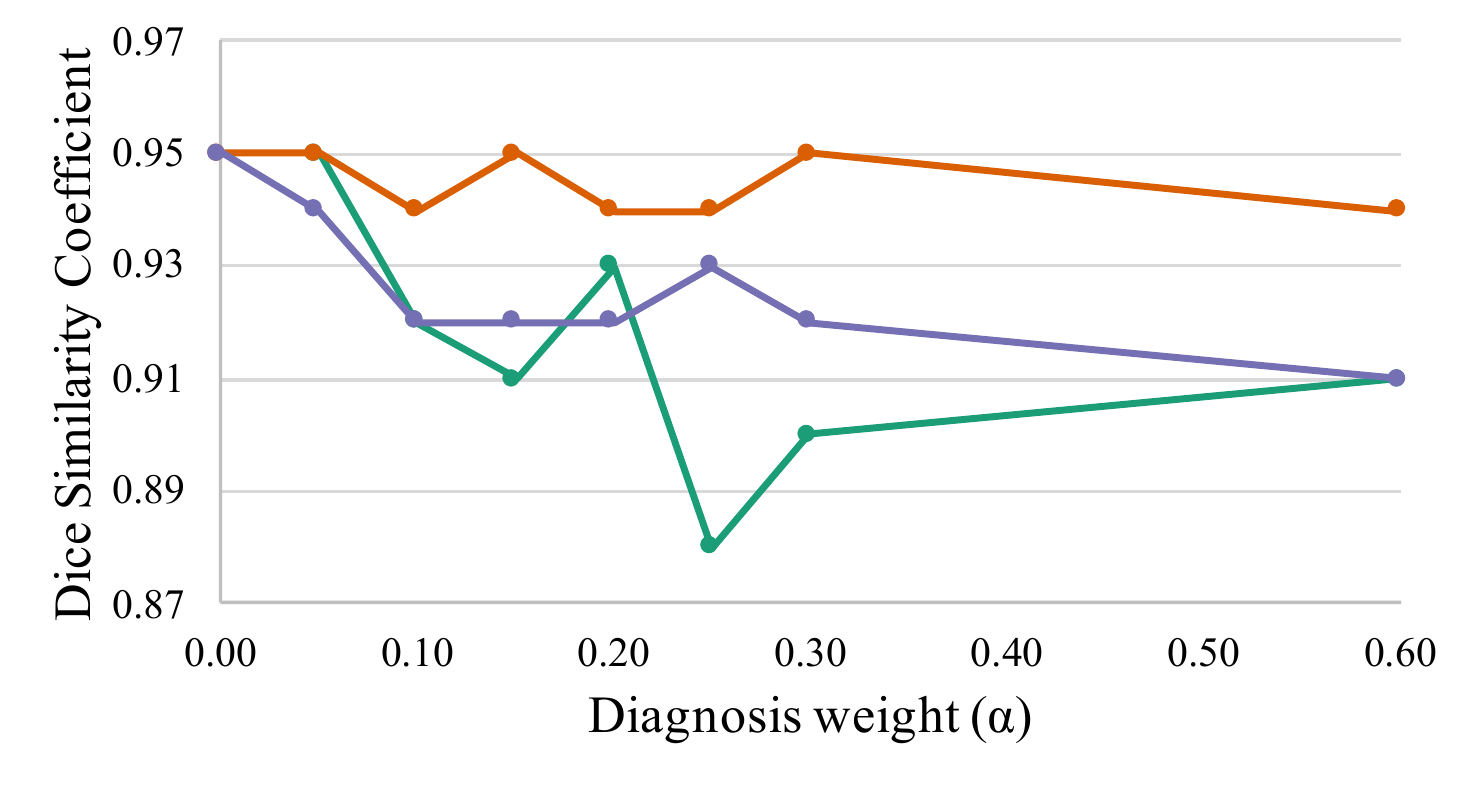}}
    \end{minipage}
    
    \begin{minipage}[b]{0.4\linewidth}
      \centering
      \centerline{\includegraphics[width=8cm]{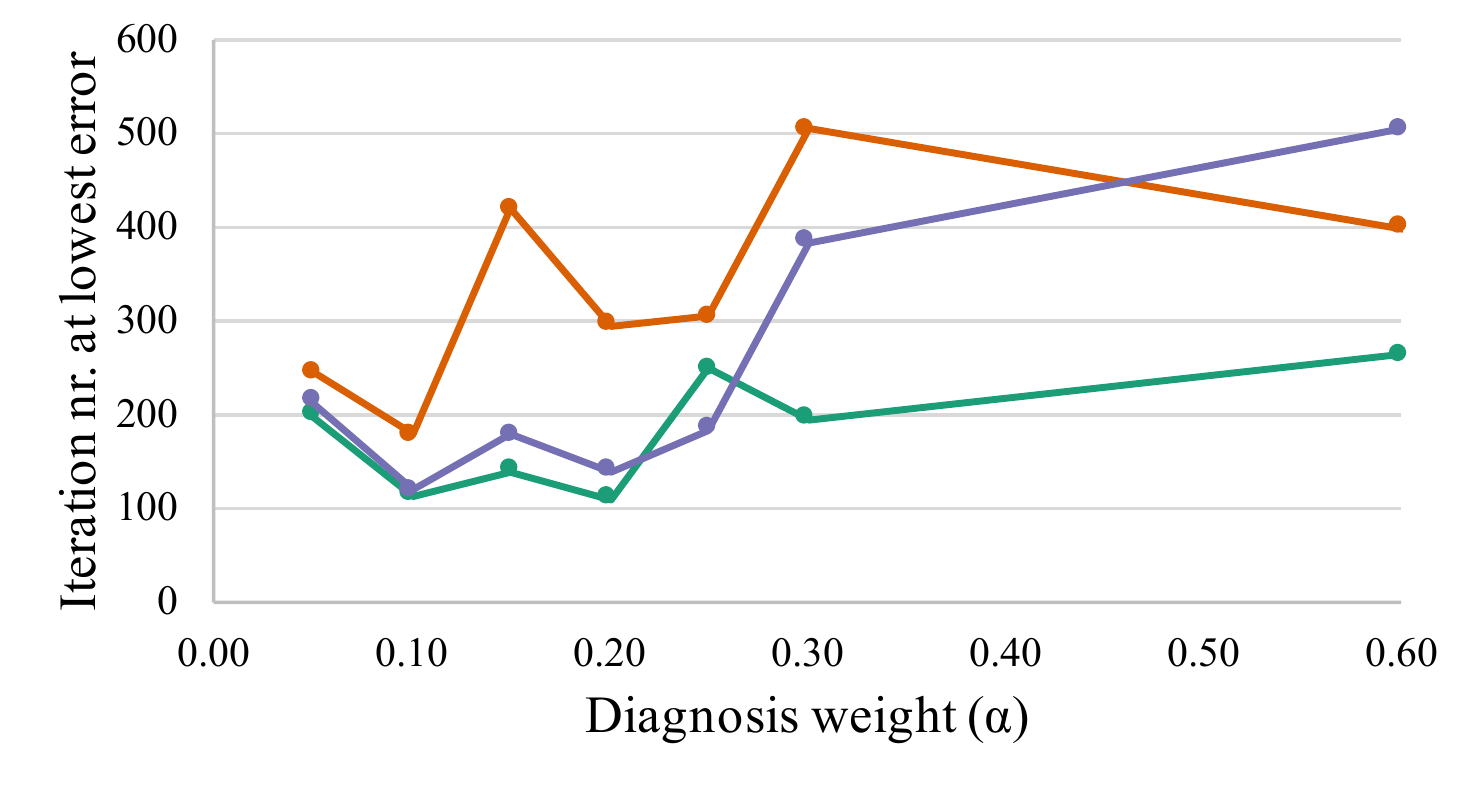}}
    \end{minipage}
    \hfill
    \begin{minipage}[b]{0.4\linewidth}
      \centering
      \centerline{\includegraphics[width=8cm]{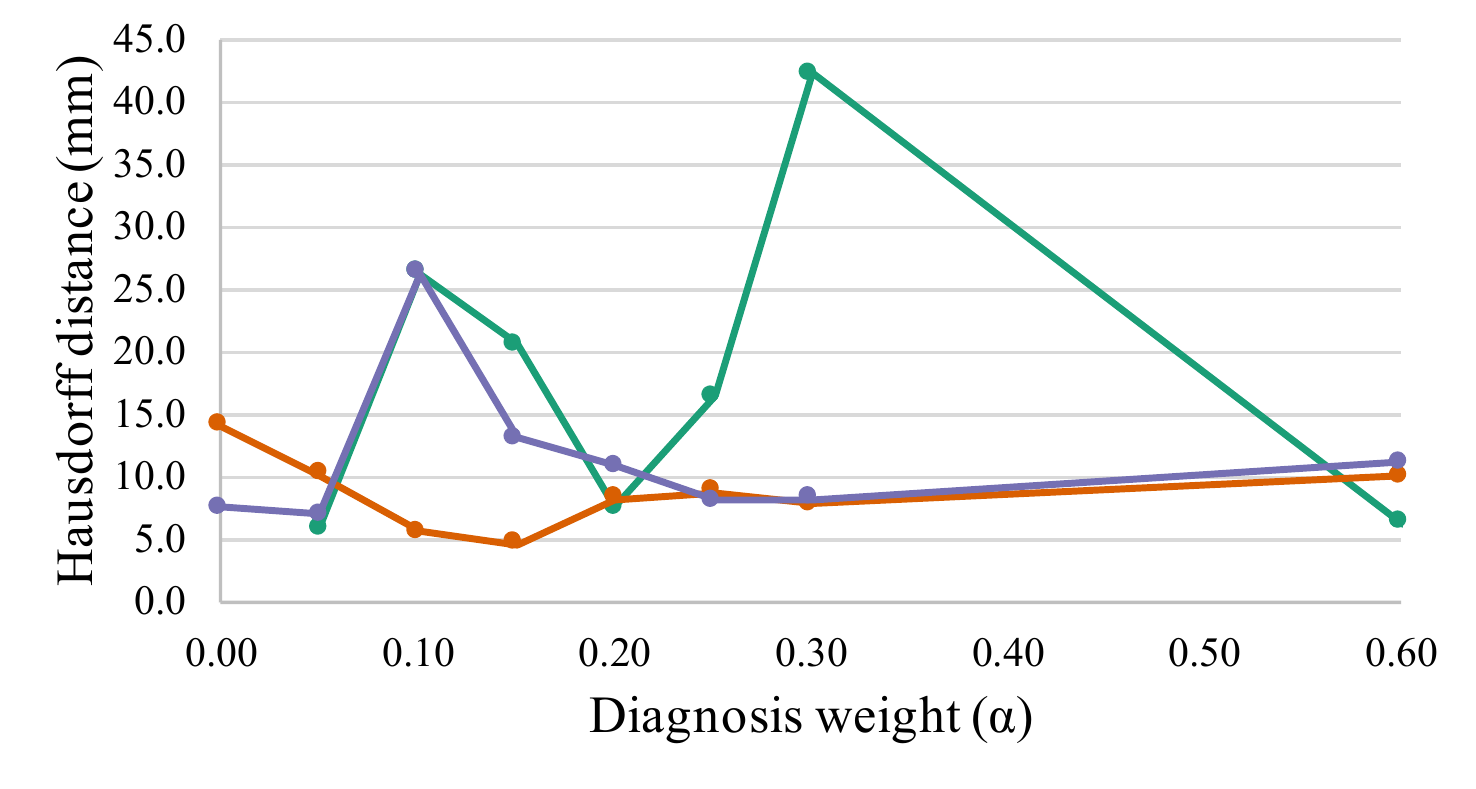}}
    \end{minipage}
    
    \caption{Results on the validation set for classification and segmentation as a function of $\alpha$ in \eqref{eq:loss} and $p$ in \eqref{eq:seg_loss}.  Left-top: diagnostic error. Left-bottom: iteration where lowest classification error happens. Right: DSC (top) and Hausdorff distance (bottom) for LV at ED.}
    \label{fig:res}
\end{figure*}

The magnitude of the gradient loss in \eqref{eq:seg_loss} increases for low Dice scores and decreases for high Dice scores if $p>1$ compared to $p=1$. This behavior emphasizes learning from cases with low Dice scores. 
For $p<1$ this emphasis is reversed, where low Dice scores are correlated with low gradient magnitude, and high Dice scores induces large gradient magnitude. The emphasis on high performing cases focuses the training process on the hard to learn details of near perfect results.
Both strategies could potentially increase segmentation performance.
In~\cite{Wong2018} an exponential logarithmic loss is evaluated that combines both strategies, but in this paper we decided to study each strategy independently.

\subsection{Training and Testing Strategies} \label{ssec:training}
The training set is split $75:25$ in equally distributed disease sets for training and validation, respectively. During each training iteration, we randomly sample six consecutive slices from the volume to be used as the input, where the center of the six slices is randomly selected.  We relied on such strategy because the dataset volumes have between 6 and 18 slices, and the normalization of this resolution (i.e., all volumes interpolated to have 6 slices) would introduce artifacts that could have a negative impact in the training and inference processes.  Additionally, such random selection of the six consecutive slices improved the training convergence and generalization.  For the inference, we use as input the center six slices of the test volume.
The Adam solver with $\beta s$ $(0.9, 0.999)$ is used for training with a learning rate of $5e^{-4}$. Also, dropout with probability $0.2$ is applied to the input layer and $0.5$ to every convolution. 

\vspace{-.05in}

\section{Results and Discussion} \label{sec:results}
\vspace{-.05in}

\begin{figure*}[t]
    \begin{minipage}[b]{0.4\linewidth}
      \centering
      \centerline{\includegraphics[width=8cm]{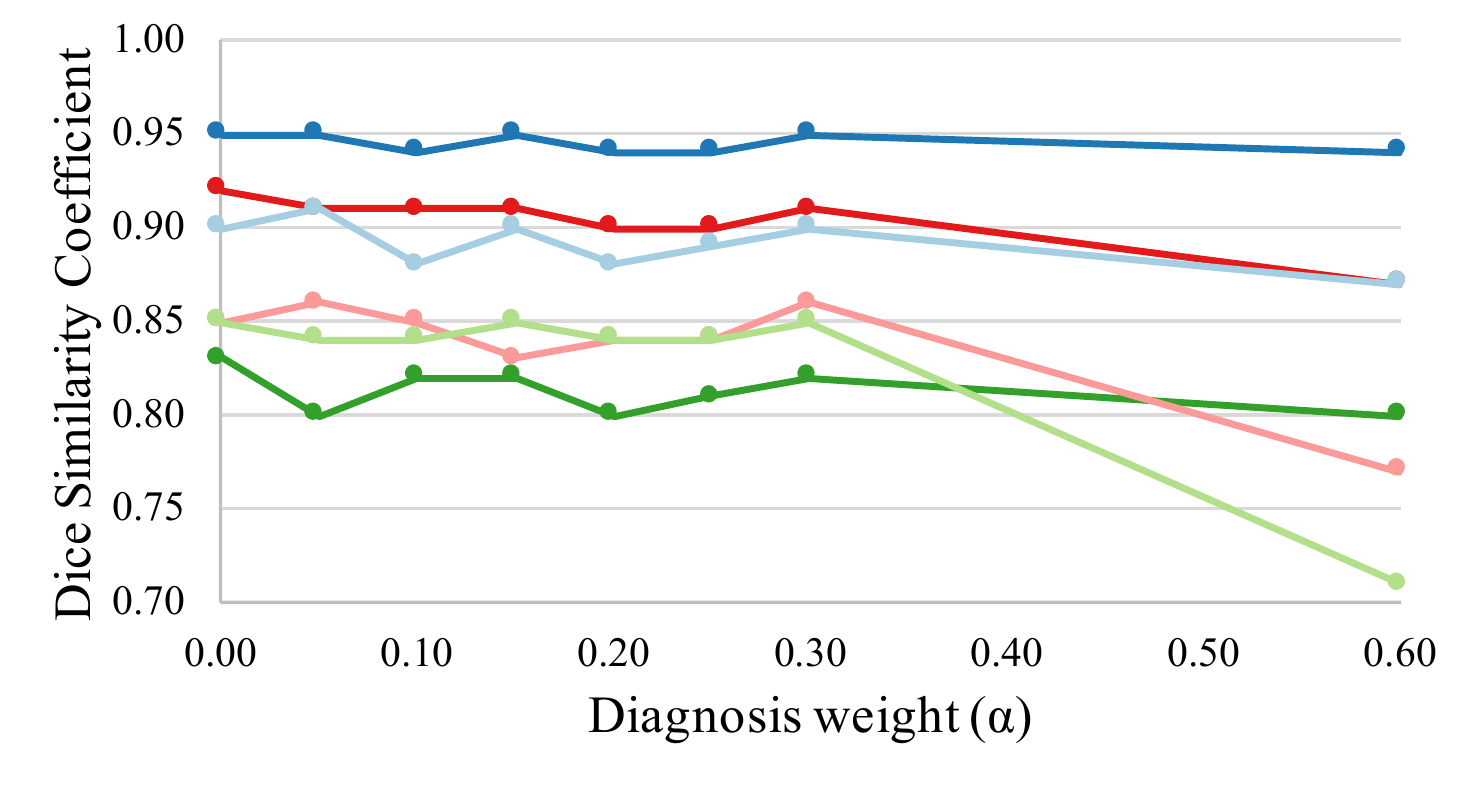}}
    \end{minipage}  
    \hfill
    \begin{minipage}[b]{0.4\linewidth}
      \centering
      \centerline{\includegraphics[width=8cm]{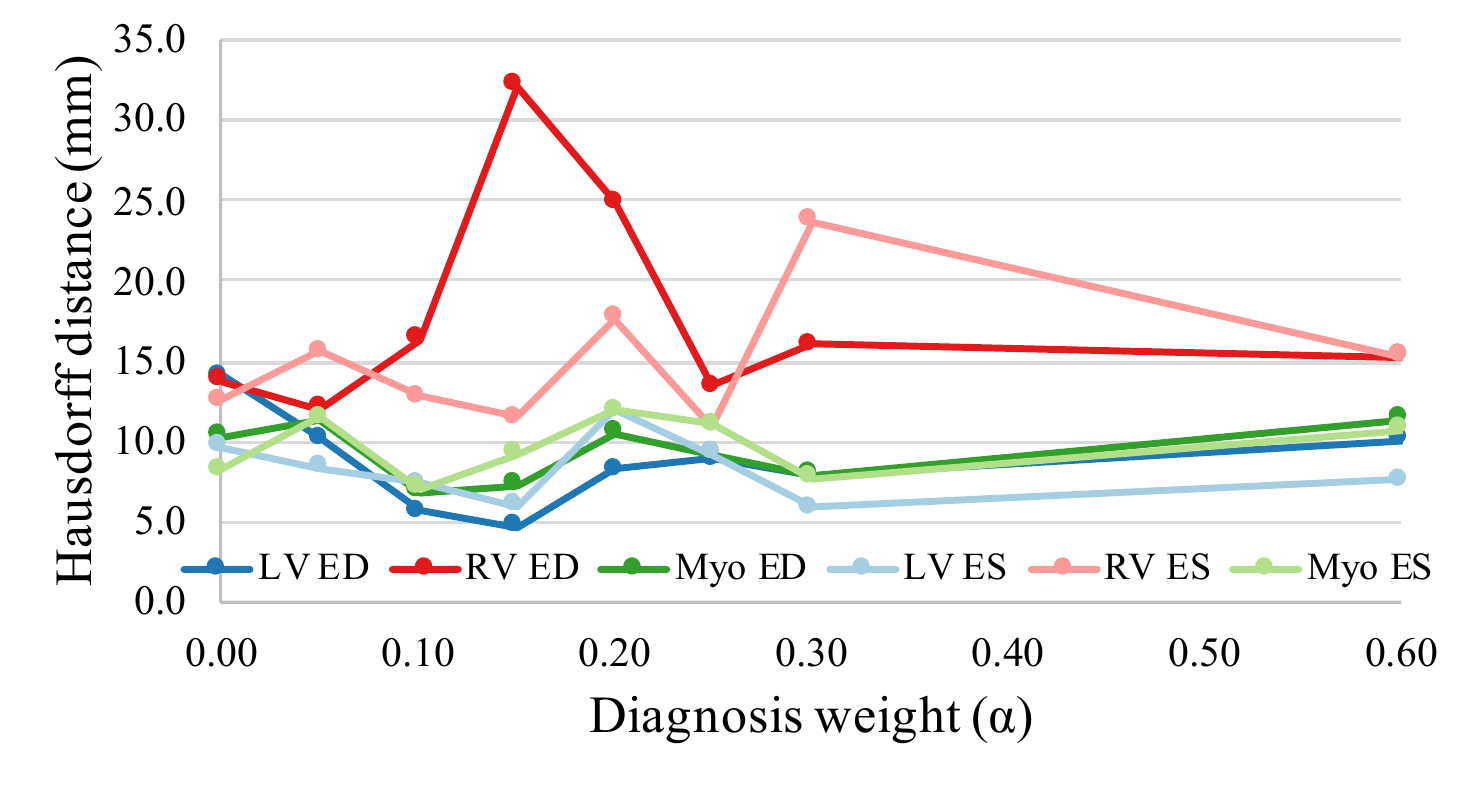}}
    \end{minipage}
        
    \begin{minipage}[b]{0.3\linewidth}
      \centering
      \centerline{\includegraphics[width=5.5cm]{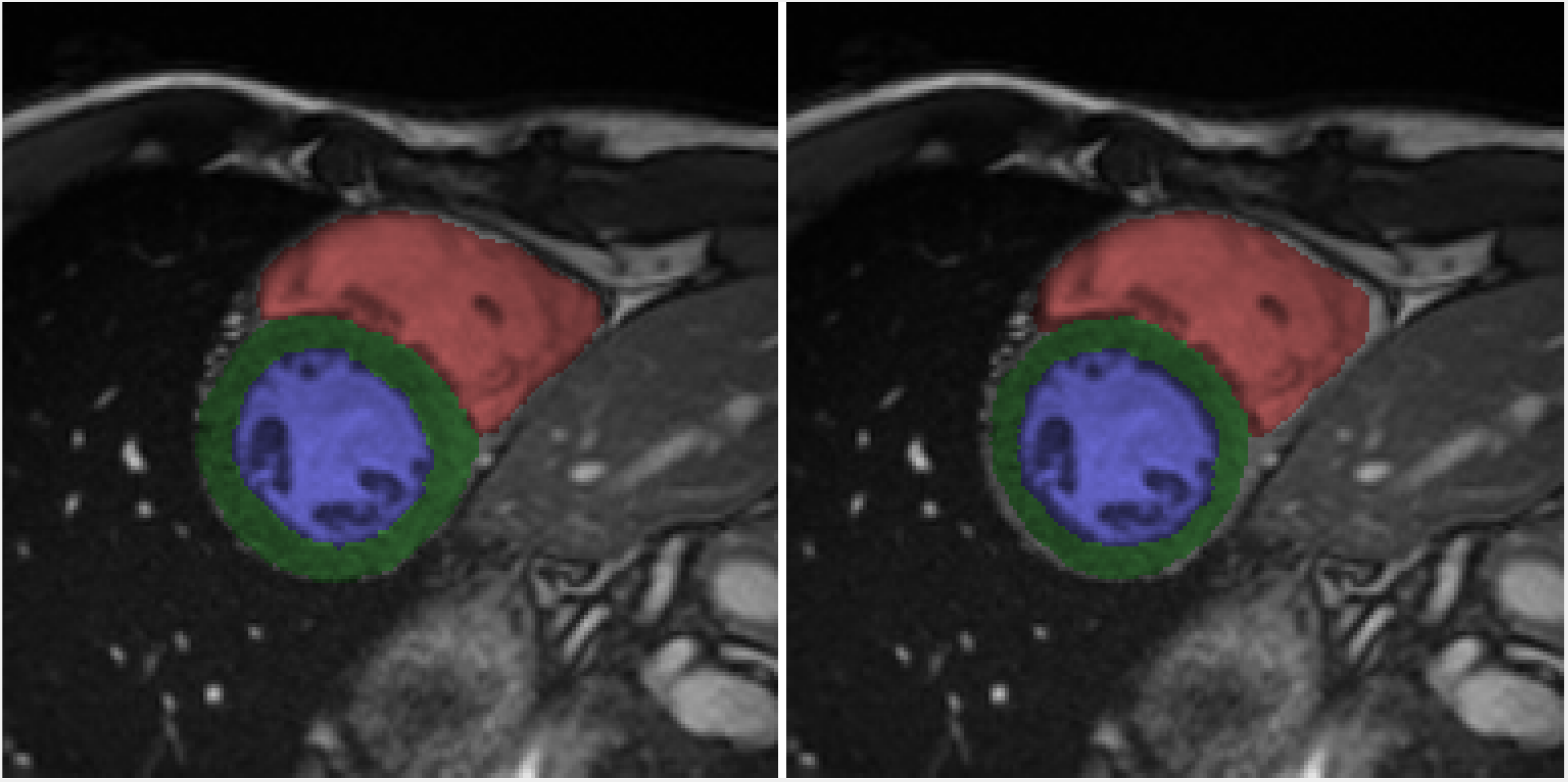}}
    \end{minipage}
    \hfill
    \begin{minipage}[b]{0.3\linewidth}
      \centering
      \centerline{\includegraphics[width=5.5cm]{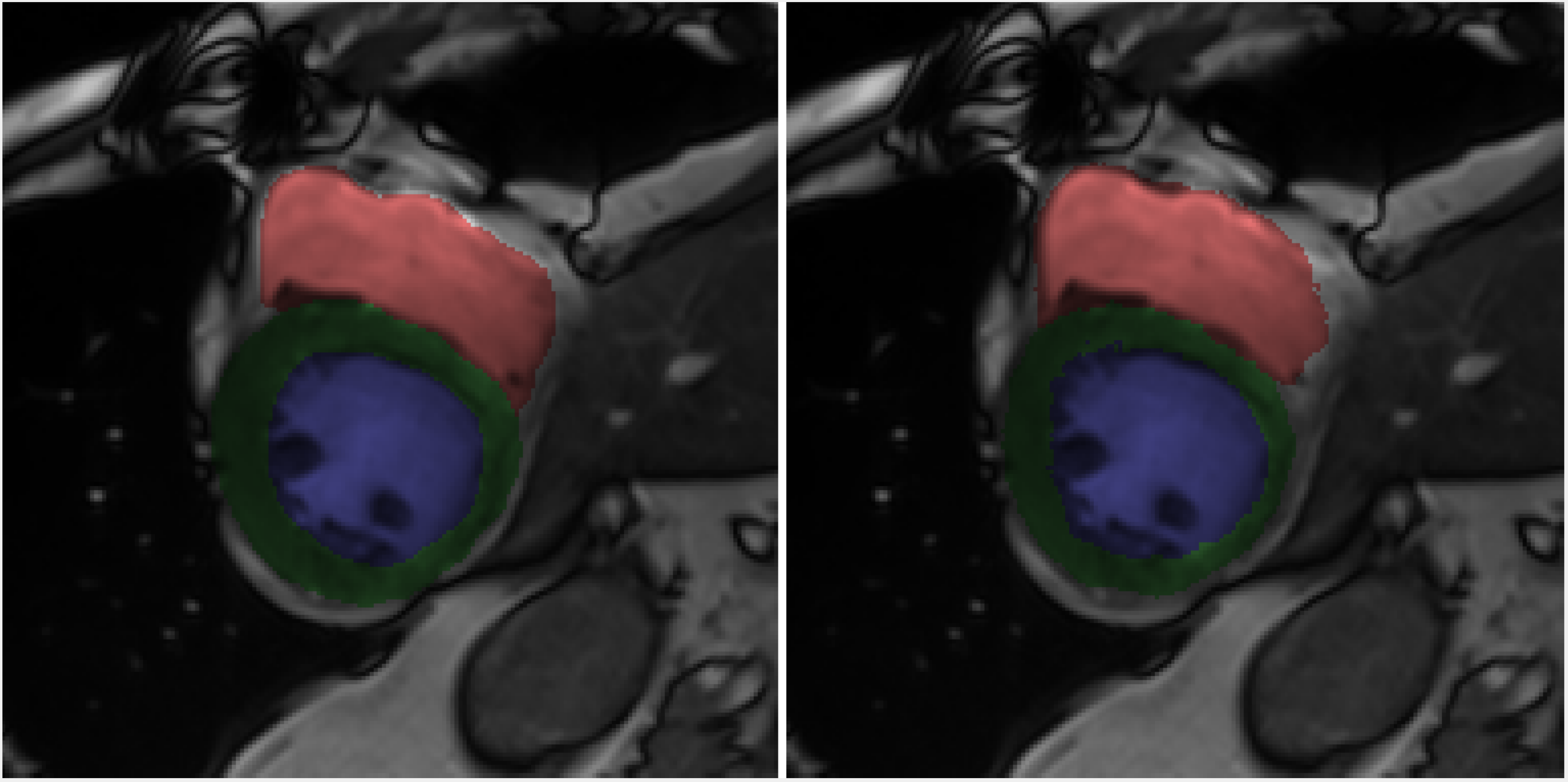}}
    \end{minipage}
    \hfill
    \begin{minipage}[b]{0.3\linewidth}
      \centering
      \centerline{\includegraphics[width=5.5cm]{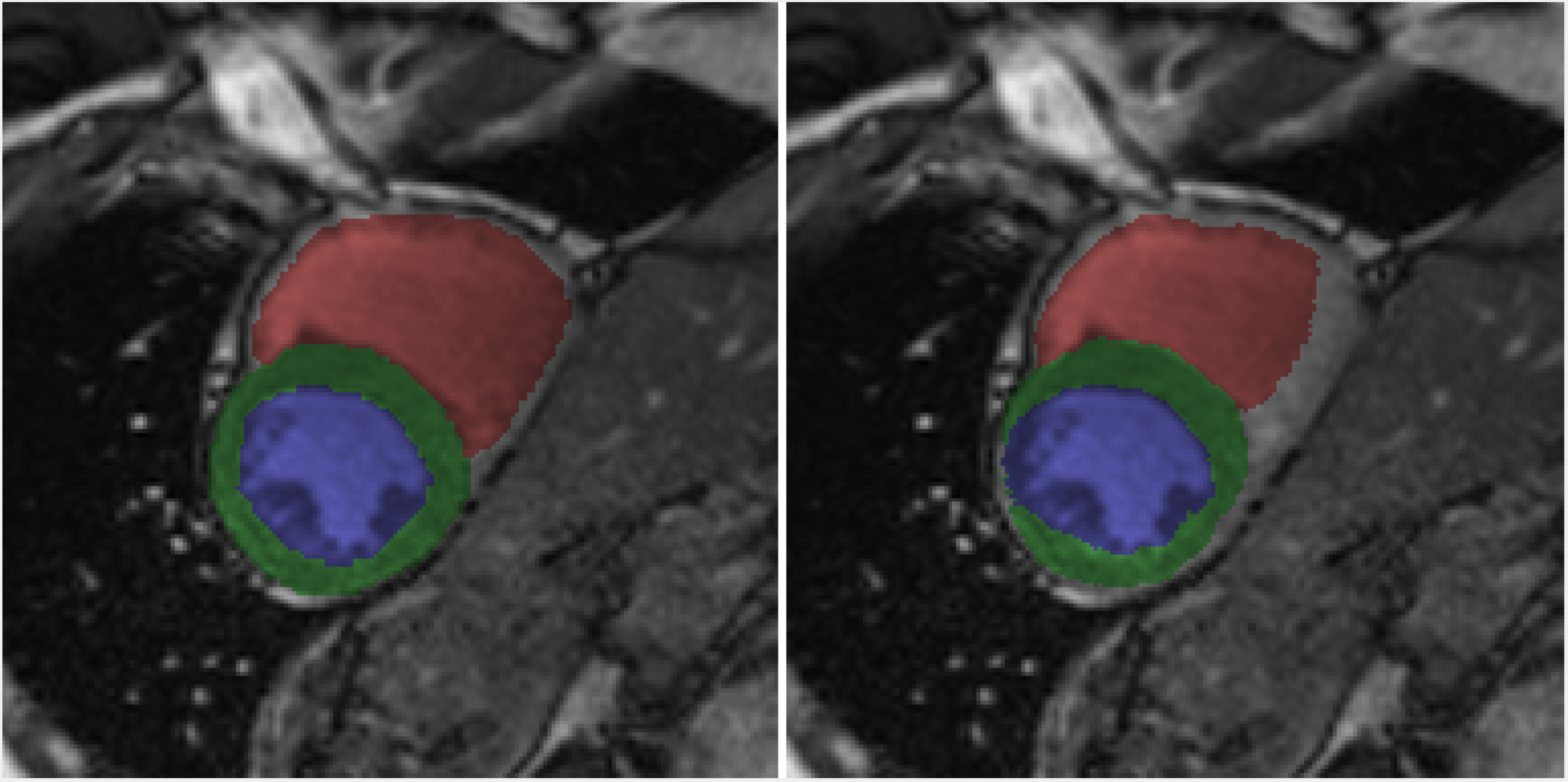}}
    \end{minipage}
    
    \caption{Segmentation results on the validation set -- Top row shows Dice Similarity Coefficients and Hausdorff distance as a function of $\alpha$ in \eqref{eq:loss} for all three segmentation anatomies and both phases for $p=0.3$.
    Bottom row shows ground truth (left) vs. segmentation results (right) for $\alpha=0.05$ and $p=0.3$. From left to right, the cases are: NOR (correcly diagnosed), MINF (correctly classified even with artifact on top left), and ARV (misclassified as NOR).
    All show {\color{blue!90!black}LV in blue}, {\color{red!90!black}RV in red}, and {\color{green!60!black}Myo in green}.
    }
    \label{fig:res_seg}
\end{figure*}

\begin{table}
    \centering
    \caption{Diagnostic accuracy on test set -- ours vs.~challenge~\cite{Bernard2018}. $^\star$~Improved accuracy to 100\% after the challenge.}
    \label{tab:accuracy}
    \begin{tabular}{lr}
        \toprule
        Model                       &Accuracy \\ \hline
        \midrule
        Baseline   ($\alpha=1.0)$   &0.68\% \\ 
        Multi-task ($\alpha=0.15)$  &0.70\% \\ 
        Multi-task ($\alpha=0.05)$  &0.78\% \\ 
        \midrule
        Wolterink et al.            &0.86\% \\ 
        Isensee et al.              &0.92\% \\ 
        Cetin et al.                &0.92\% \\ 
        Khened et al.$^\star$       &0.96\% \\ 
        \bottomrule
    \end{tabular}
\end{table}

In this section, for the classification results, we rely on the rate of diagnostic error~\cite{Bernard2018}, while the segmentation results rely on Dice similarity coefficient (DSC) and Hausdorff distance~\cite{Bernard2018}.  We first study the influence of $\alpha \in [0.0,0.6]$ in~\eqref{eq:loss} and $p\in\{0.3,1,2\}$ in~\eqref{eq:seg_loss}. Results in Fig.~\ref{fig:res} show that a low value of $\alpha$ corresponds to lower diagnosis classification error, faster convergence, and better (i.e., higher) Dice scores. 
In fact, the best result with $p=0.3$ for $\alpha=1$ (i.e., this represents an optimization that consists of only the diagnosis loss) is reached at iteration 680 (not show in figure), while for $\alpha=0.05$, the best result is reached $2.5\times$ faster.
The Hausdorff distance is the only metric where performance seems to decrease with a lower value for $\alpha$, however, this could be due to the  sensitivity of this metric to outliers rather than an actual influence of $\alpha$.
All three values of $p$ seem to produce similar diagnostic error, and convergence to the lowest diagnosis classification error is faster when $p \geq 1$, but $p = 0.3$ shows superior DSC and Hausdorff distance results, compared to $p \geq 1$. 
Contrary to the other values for $p$, results indicate minimal or no influence of the diagnosis loss on the  segmentation results for $p=0.3$. This shows that the large gradient magnitude in the segmentation loss for well performing cases overwhelms the influence of classification training, making segmentation performance independent of diagnosis training for lower values of $\alpha$. For $p=2.0$ on the other hand, low gradient magnitude at high DSC causes the model to focus on the diagnosis loss, reducing segmentation accuracy as $\alpha$ increases. The robust diagnosis and segmentation results achieved with $p=0.3$ makes this model best suited for our multi-task training approach.

Fig.~\ref{fig:res_seg} shows consistent accuracy for the segmentation of all anatomies (\{LV,RV,Myo\} at \{ED,ES\}) over a large range of values for $\alpha$.  In terms of DSC, Myo is the hardest anatomy to segment, while regarding Hausdorff distance, RV appears to be challenging because of the sensitivity of this distance measure to outliers.  These results are around $2\%$ to $7\%$ worse than the best ones in the ACDC challenge~\cite{Bernard2018}. 
However, no direct comparison can be made with the ACDC challenge results~\cite{Bernard2018} as we did not obtain results on the test set because of the inference approach described in Sec.~\ref{ssec:training} consisting of the assessment of the six central slices per volume.

In diagnosis classification, we evaluate three values of $\alpha\in\{0.05,0.15,1\}$ on the test set of the ACDC challenge~\cite{Bernard2018} (with $p=0.3$ in Eq.~\eqref{eq:seg_loss}). Table~\ref{tab:accuracy} shows an increase in accuracy when $\alpha$ decreases, which is consistent with the observations in Fig.~\ref{fig:res} on the validation set.
The decrease in accuracy from $\alpha=0.05$ to $\alpha=0.15$ can in part be explained by their difference in segmentation performance. Of the fifteen misclassifications for $\alpha=0.15$, seven involve the ARV class. This coincides with the observed sharp increase of the RV Hausdorff distance around $\alpha=0.15$ in Fig.~\ref{fig:res_seg}. DSC and Hausdorff distances score well for $\alpha=0.05$ and no clear observations can be made about the origin of misclassifications without looking at the images. As the test server provides no information on individual cases, we perform further evaluation on the validation set.

The bottom row of Fig.~\ref{fig:res_seg} shows the  segmentation results along with their ground truth for $p=0.3$ and $\alpha=0.05$, where three out of twenty-five cases are misdiagnosed (we show two correct and one incorrect diagnosed case). 
For the case that has been incorrectly diagnosed in Fig.~\ref{fig:res_seg} (rightmost image, ARV misclassified as NOR), the shown under-segmented RV segmentation is representative of the entire ES phase, while for the ED phase, the RV is correctly segmented.  Given that ARV relies on RV ejection fraction, such mistake in the segmentation would suggest 
adequate myocardial contraction and explain why this case is classified as normal. 
This mistake provides evidence that the features used for training the classification parameters may be strongly correlated with segmentation accuracy. 
The other two misclassifications involve scans that have imaging artifacts near the heart, similar to the middle image in the bottom row of Fig.~\ref{fig:res_seg}. Interestingly, the only four cases in the validation set that contain imaging artifacts have a softmax probability (i.e., classification confidence) of around 0.7, while all other scans have a probability near 1. Segmentation performance is unaffected by such artifacts, and two of these four cases are still correctly classified, but it does show that classification performance suffers in scans with imaging artifacts.

Comparing our results to the ACDC challenge results (Table~\ref{tab:accuracy}) that used handcrafted features for diagnosis, we see that the accuracy of our model needs to be improved by $10$ to $20\%$ to become competitive. This was expected as the handcrafted features used by the state-of-the-art methods are the same as those used in clinical diagnosis. 
However, our model shows promising results and could possibly reach similar performance when larger datasets become available. 
Furthermore, the ACDC challenge organizers excluded ambiguous cases that contained  diagnostic boundary values for the handcrafted features. 
This design choice provided large margins for the classifiers to place their decision boundaries on when using handcrafted features. It is likely that adding these clinical boundary cases will have a significant impact on the performance of such methods.

\vspace{-.05in}

\section{Conclusion} 
\label{sec:conclusion}
\vspace{-.05in}

In this paper, we show the first competitive end-to-end diagnosis and segmentation training from CMR imaging. We show that multi-task training can converge $2.5\times$ faster and reduce the diagnostic error from $32\%$ to $22\%$ compared to a baseline method trained without segmentation. To the best of our knowledge, this is the best result of an end-to-end segmentation and classification method for diagnosing from CMR.
Nevertheless, our results need to be improved further before they become competitive with state-of-the-art methods that rely on handcrafted features. 
We believe that his is simply a matter of increasing the dataset used for training, so we plan to focus on this issue as our future research activity.

\bibliographystyle{IEEEbib}
\bibliography{refs}

\end{document}